\documentclass[conference]{IEEEtran}
\IEEEoverridecommandlockouts
\usepackage{cite}
\usepackage{amsmath,amssymb,amsfonts}
\usepackage{amsthm}
\usepackage{algorithmic}
\usepackage{graphicx}
\usepackage{textcomp}
\usepackage{xcolor}
\usepackage{booktabs}
\usepackage{multirow}
\usepackage{array}
\def\BibTeX{{\rm B\kern-.05em{\sc i\kern-.025em b}\kern-.08em
    T\kern-.1667em\lower.7ex\hbox{E}\kern-.125emX}}
\begin{document}
\title{IPM-FM: A Foundation Model with Consensus Feature Selection for Industrial Process Monitoring}

\author{
\IEEEauthorblockN{Liang Cao}
\IEEEauthorblockA{Department of Chemical and Biological Engineering\\
University of British Columbia\\
Vancouver, BC, Canada}
\and

\IEEEauthorblockN{Weide Liu}
\IEEEauthorblockA{School of Computing and Artificial Intelligence\\
Jiangxi University of Finance and Economics\\
Nanchang, China}

\and
\IEEEauthorblockN{Yan Qin}
\IEEEauthorblockA{School of Automation\\
Chongqing University\\
Chongqing, China}

\and
\IEEEauthorblockN{Jun Cheng}
\IEEEauthorblockA{Institute for Infocomm Research\\
A*STAR\\
Singapore}

\and
\IEEEauthorblockN{Weisi Lin}
\IEEEauthorblockA{College of Computing and Data Science\\
Nanyang Technological University\\
Singapore}

\and
\IEEEauthorblockN{Bhushan Gopaluni}
\IEEEauthorblockA{Department of Chemical and Biological Engineering\\
University of British Columbia\\
Vancouver, BC, Canada}
}

\maketitle

\begin{abstract}
Industrial process monitoring is fundamental to the safety and economic performance of modern process plants. Current practice remains a one-task-one-model paradigm that is label-inefficient and prone to degradation under operating drift. Foundation models have reshaped language, vision, and generic time-series forecasting, but it has not been adapted to industrial process monitoring. This setting poses domain-specific challenges, including safety-critical decisions and asymmetric sampling between process variables and laboratory measurements. We propose the industrial process monitoring foundation model (IPM-FM). It first learns general-purpose representations from unlabeled industrial process data through self-supervised pretraining, then adapts to specific monitoring tasks using a small amount of task-labeled data, and finally produces calibrated predictions through an uncertainty-aware prediction head. IPM-FM integrates a self-supervised Informer backbone with a multi-criteria consensus feature selector, a recursive lag-feature regression head, and a calibrated Monte Carlo dropout uncertainty module. On a seven-year hydrotreater dataset for diesel flash-point soft sensing, IPM-FM attains an RMSE of 2.99, $R^2$ of 0.50, and 97\% coverage of its 95\% predictive interval, outperforming the strongest classical and from-scratch sequence baselines by 8.3\% and 14.6\% in RMSE respectively, supporting the viability of a unified pretraining--adaptation framework for industrial process monitoring.
\end{abstract}

\begin{IEEEkeywords}
Foundation model, industrial process monitoring, soft sensing, self-supervised pretraining, causal feature selection, uncertainty quantification.
\end{IEEEkeywords}

\section{Introduction}
Modern industrial plants increasingly rely on dense sensor networks~\cite{yin2014review}. Soft sensing infers hard-to-measure quality variables from cheap online measurements~\cite{kadlec2009review,cao2025comprehensive}. Recent automated soft-sensor design tools have further highlighted the need for deployable machine-learning workflows in industrial applications~\cite{cao2025automatedsoftsensor}. Fault detection flags significant deviations from a learned normal operating region. Fault diagnosis isolates the root cause among known failure modes, while prognostic methods estimate the remaining useful life of critical equipment. Although these tasks consume the same multivariate process time series, current practice addresses each one with a dedicated model trained from scratch on its own labeled history.

This one-task-one-model paradigm carries three fundamental limitations. First, it is data inefficient because task labels are scarce and expensive while modern neural models are label-hungry. Second, it offers little transferability because knowledge extracted from one task or plant is rarely reused, even though industrial signals share common dynamical primitives such as oscillations, step responses, and slow drifts. Third, it degrades quickly under operating drift because models fitted to a specific operating window fail when feedstock, catalyst activity, or ambient conditions shift, and each drift then forces an expensive retraining cycle on every affected task. This issue is very common in multimode industrial processes, where adaptive monitoring across operating modes is required~\cite{cao2025adaptiveprocessmonitoring}. Addressing these issues requires decoupling representation learning from task supervision so that a single learned representation can serve many monitoring tasks.

Such decoupling is the contribution of foundation models in language and vision, where a single backbone is pretrained on massive unlabeled data with self-supervised objectives and then adapted to heterogeneous downstream tasks with minimal labels~\cite{devlin2019bert,he2022masked}. The same framework has recently been extended to generic time-series data through large pretrained backbones, with notable examples including Chronos~\cite{ansari2024chronos} and TimesFM~\cite{das2024timesfm} for zero-shot forecasting and Moirai~\cite{woo2024moirai} and MOMENT~\cite{goswami2024moment} for general-purpose representation. Earlier representation learners such as TS2Vec~\cite{yue2022ts2vec} together with task-specific backbones such as PatchTST~\cite{nie2023time} and TimesNet~\cite{wu2023timesnet} have further pushed the state of the art on standard forecasting and classification benchmarks. 

Complementary progress in geometric representation learning has produced important methodological advances for extracting structured and interpretable primitives from complex physical observations. In particular, BPNet and its extension provide an advanced framework for B{\'e}zier primitive segmentation and decomposition directly from irregular 3D point clouds, demonstrating that deep networks can learn compact, primitive-level abstractions from unstructured geometric data~\cite{fu2023bpnet,fu2025bezierdecomposition}. Nevertheless, these advances are designed for generic time-series forecasting or geometric data abstraction rather than industrial process monitoring. They treat data either as generic numerical sequences or as geometric point sets, rather than as signals from physically coupled actuators and sensors inside a controlled plant.

\begin{figure*}[t]
\centering
\includegraphics[width=0.98\textwidth]{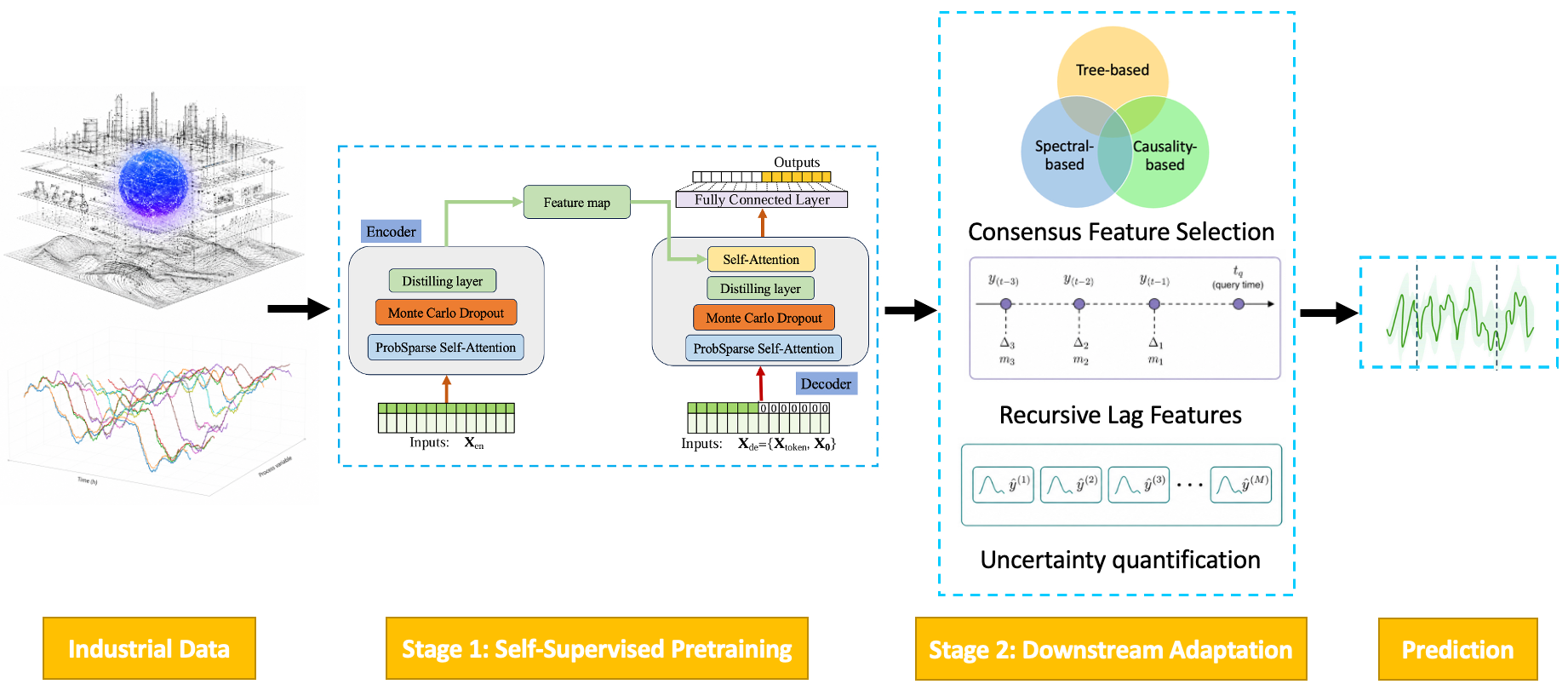}
\caption{Two-stage IPM-FM pipeline: self-supervised pretraining of a shared backbone, followed by consensus feature selection, recursive lag features, and a calibrated uncertainty head for downstream adaptation.}
\label{fig:framework}
\end{figure*}

Transferring the foundation-model to industrial process monitoring therefore faces three domain-specific challenges. First, industrial plants carry dozens to hundreds of sensor channels, only a task-dependent subset of which carries useful signal. Spurious channels become a dominant failure mode whenever raw multivariate input is fed directly into a shared backbone. Second, monitoring decisions are safety-critical and calibrated uncertainty becomes a hard requirement rather than an optional feature~\cite{gawlikowski2023survey,guo2017calibration}. Third, the sampling rate between process variables and laboratory measurements is severely mismatched. Laboratory information must therefore be effectively exploited to keep the downstream model up to date.

Motivated by these gaps, we propose industrial process monitoring foundation model (IPM-FM), a model that addresses all three challenges within a single unified architecture. We validated the framework for diesel flash-point soft sensing at a commercial hydrotreating unit. IPM-FM outperforms the strongest classical and from-scratch sequence baselines by 8.3\% and 14.6\% RMSE respectively.  The main contributions of this paper are summarized as follows:
\begin{itemize}
    \item We propose IPM-FM, the first foundation-model framework tailored to industrial process monitoring, which decouples representation learning from task supervision and unifies multiple monitoring tasks under a shared self-supervised backbone.
    \item We design a multi-criteria consensus feature selector that fuses tree-based, spectral, and causal evidence, with a theoretical guarantee of multiplicative suppression of spurious channels.
    \item We introduce a recursive lag-feature mechanism that exploits sparse but highly informative laboratory measurements, together with a calibrated MC dropout wrapper that delivers predictive intervals suitable for risk-aware deployment.
\end{itemize}

\section{IPM-FM Framework}
\label{sec:framework}
Figure~\ref{fig:framework} illustrates the overall architecture of IPM-FM, which is organized into a two-stage pipeline. In the pretraining stage, a shared backbone is trained on unlabeled multivariate process data using self-supervised objectives that do not require any monitoring labels. In the adaptation stage, the pretrained backbone is combined with a consensus feature selection module, a recursive lag-feature builder, and a calibrated uncertainty head to produce a downstream monitor for a specific task. Switching among monitoring tasks is achieved by replacing only the lightweight task head while preserving the pretrained backbone and the consensus feature adaptation module, which yields the cross-task reusability characteristic of foundation-model pipelines.

\subsection{Notation and Problem Setup}
Let $\mathbf{X} \in \mathbb{R}^{T \times D}$ denote a segment of multivariate process data with $T$ time steps and $D$ sensor channels. Different downstream monitoring tasks correspond to different target spaces: for soft sensing, a scalar quality variable $y \in \mathbb{R}$ is associated with the final time step; for fault detection and diagnosis, a categorical label $c \in \{1,\ldots,C\}$ is attached instead; for anomaly detection, no target label is required at inference. The pretraining set $\mathcal{D}_{\text{pre}} = \{\mathbf{X}^{(i)}\}_{i=1}^{N_{\text{pre}}}$ consists of unlabeled segments, while the downstream set $\mathcal{D}_{\text{down}}$ contains aligned input-target pairs for the task of interest, with $|\mathcal{D}_{\text{down}}| \ll N_{\text{pre}}$. The goal is to learn a backbone $g_\phi$ pretrained on $\mathcal{D}_{\text{pre}}$ and a lightweight task head $h_\psi$ such that $h_\psi(g_\phi(\mathbf{X}), \mathbf{l})$ approximates the downstream target with calibrated uncertainty, where $\mathbf{l}$ collects the most recent laboratory measurements. In this paper we instantiate the regression setting for soft sensing, writing $\hat{y} = f_\theta(\mathbf{X}, \mathbf{l})$ with $\theta = (\phi,\psi)$; the generalization to classification and anomaly heads is straightforward and is treated as future work.

\section{Self-Supervised Pretraining}
\label{sec:pretraining}

\subsection{Backbone Architecture}
\begin{figure}[t]
\centering
\includegraphics[width=0.95\columnwidth]{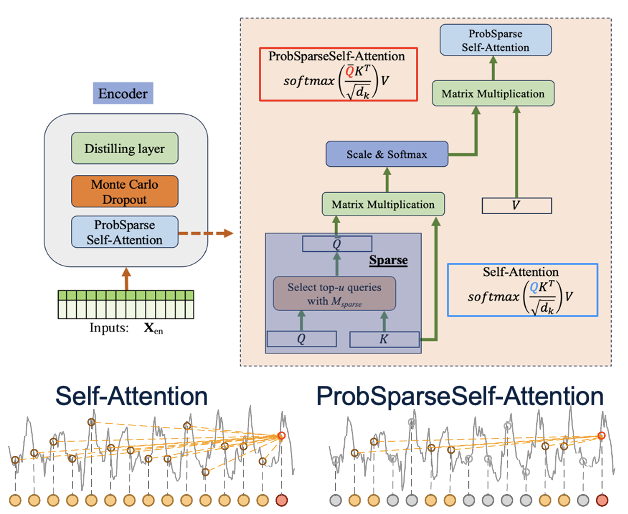}
\caption{Informer-based encoder backbone of IPM-FM, with stacked ProbSparse self-attention and distillation layers for long industrial sequences.}
\label{fig:probsparse}
\end{figure}

The backbone is an encoder architecture derived from the Informer~\cite{zhou2021informer}, which itself follows the standard transformer attention construction~\cite{vaswani2017attention}, chosen for its favorable trade-off between long-range modeling capacity and computational cost on industrial sequences. Informer-based architectures have also shown promise for interpretable industrial soft-sensor design with long process sequences~\cite{cao2024informerShap}. The encoder ingests a segment $\mathbf{X} \in \mathbb{R}^{T \times D}$ and produces a sequence of latent representations $\mathbf{H} \in \mathbb{R}^{T' \times d_{\text{model}}}$ through stacked ProbSparse self-attention and distillation layers. Figure~\ref{fig:probsparse} shows the overall encoder structure used as the IPM-FM backbone. The same backbone is shared across all downstream monitoring tasks.

\subsubsection{ProbSparse Self-Attention}
Standard self-attention computes:
\begin{equation}
\text{Attn}(\mathbf{Q},\mathbf{K},\mathbf{V}) = \text{softmax}\!\left(\frac{\mathbf{Q}\mathbf{K}^\top}{\sqrt{d_k}}\right)\mathbf{V},
\end{equation}
which scales quadratically with sequence length. To reduce the cost on long industrial sequences, the ProbSparse mechanism restricts the interaction to the top-$u$ queries selected according to a sparsity measure:
\begin{equation}
M_{\text{sp}}(\mathbf{q}_i) = \max_j \mathbf{A}_{ij} - \frac{1}{n}\sum_{j=1}^{n}\mathbf{A}_{ij},
\end{equation}
where $\mathbf{A} = \mathbf{Q}\mathbf{K}^\top / \sqrt{d_k}$, $n$ is the sequence length, and $u = \lceil \kappa \log n \rceil$ with $\kappa = 3$. Only a sparse subset of dominant query--key interactions is retained, which sharply reduces memory and FLOPs on industrial sequences with hundreds to thousands of time steps.

\subsubsection{Distillation Layer}
To compress the sequence length and reduce memory consumption, each encoder stage concludes with a distillation layer
\begin{equation}
\mathbf{X}_{\text{out}} = \text{MaxPool}\big(\text{ELU}(\mathbf{W}*\mathbf{X}_{\text{in}} + \mathbf{b})\big),
\end{equation}
which halves the sequence length while preserving the most informative activations. 

\subsection{Self-Supervised Objectives}
Pretraining is driven by two complementary objectives that do not require any task labels and therefore produce a representation that is reusable across monitoring tasks.

\subsubsection{Masked Segment Reconstruction}
Given an input segment $\mathbf{X}$, we randomly mask a fraction $r$ of non-overlapping sub-segments of length $\ell$ and task a lightweight reconstruction head to predict the masked values from the surrounding context:
\begin{equation}
\mathcal{L}_{\text{rec}} = \frac{1}{|\mathcal{M}|}\sum_{(t,d)\in\mathcal{M}} \big(\hat{\mathbf{X}}_{t,d} - \mathbf{X}_{t,d}\big)^2,
\end{equation}
where $\mathcal{M}$ is the index set of masked positions.

\subsubsection{Operating-Regime Contrastive Learning}
Industrial processes typically operate in a small number of recurrent regimes separated by transitions, and the ability to distinguish regimes is valuable for every downstream monitoring task. We exploit this structure with a contrastive objective that pulls together representations of segments from the same operating regime and pushes apart representations from different regimes. Regimes are obtained by applying $k$-means clustering to daily aggregated process statistics, which is a lightweight pseudo-labeling step that does not rely on any downstream label. The information noise-contrastive estimation loss is:
\begin{equation}
\mathcal{L}_{\text{cont}} = -\frac{1}{N}\sum_{i=1}^{N}\log\frac{\exp(\text{sim}(\mathbf{z}_i,\mathbf{z}_i^+)/\tau)}{\sum_{k\ne i}\exp(\text{sim}(\mathbf{z}_i,\mathbf{z}_k)/\tau)},
\end{equation}
where $\mathbf{z}_i$ is the pooled representation of segment $i$, $\mathbf{z}_i^+$ is a positive segment from the same regime, and $\tau$ is a temperature hyperparameter.

\subsubsection{Total Pretraining Loss}
The total pretraining loss is a weighted sum
\begin{equation}
\mathcal{L}_{\text{pre}} = \mathcal{L}_{\text{rec}} + \lambda_{\text{cont}} \mathcal{L}_{\text{cont}},
\end{equation}
where $\lambda_{\text{cont}}$ is selected on a held-out validation split.

\section{Downstream Task Adaptation}
\label{sec:adaptation}

During downstream adaptation, the pretrained backbone is connected to three task-specific modules. A consensus feature selection module identifies the subset of sensor channels relevant to the target monitoring task. A recursive lag-feature builder augments the encoder embedding with prior laboratory measurements, and a calibrated uncertainty head produces predictive intervals or class probabilities depending on the task.

\subsection{Multi-Criteria Consensus Feature Selection}
Industrial plants carry dozens to hundreds of sensor channels, many of which are irrelevant or redundant for any specific monitoring target. We propose a consensus framework that fuses three complementary families of feature selection methods.

\subsubsection{Tree-Based Importance}
Tree-based ensembles including LightGBM~\cite{ke2017lightgbm}, Extra Trees~\cite{geurts2006extra}, Random Forest~\cite{breiman2001random}, and CatBoost~\cite{prokhorenkova2018catboost} rank features by impurity decrease:
\begin{equation}
I_j = \sum_{t\in\text{Trees}} \sum_{s \in S_j^t} p(s)\cdot \Delta s,
\end{equation}
where $S_j^t$ is the set of nodes split on feature $j$ in tree $t$, $p(s)$ is the fraction of samples reaching node $s$, and $\Delta s$ is the impurity decrease.

\subsubsection{Spectral Similarity}
Power spectral densities are computed via Welch's method, and the similarity between feature $i$ and the monitoring target $y$ is scored as
\begin{equation}
\rho_{\text{spec}}(i) = \frac{\sum_f (P_i(f)-\bar P_i)(P_y(f)-\bar P_y)}{\sqrt{\sum_f(P_i(f)-\bar P_i)^2 \sum_f(P_y(f)-\bar P_y)^2}}.
\end{equation}
This criterion captures dynamic similarity that is invariant to time delays and phase shifts.

\subsubsection{Causal Discovery}
Three causal discovery algorithms are applied: the PC algorithm for constraint-based conditional independence testing, FCI for handling latent confounders, and DirectLiNGAM for exploiting non-Gaussianity~\cite{vowels2022dags, shimizu2011directlingam}. Causality analysis has been used in industrial inferential sensing and stable soft-sensor modeling to improve feature relevance, interpretability, and robustness under changing operating conditions~\cite{cao2020dynamicinferential,yu2022stablesoftsensor}. Process-knowledge-guided causal discovery can further reduce incorrect causal relations when applying causal discovery algorithms to industrial process data~\cite{cao2022causaldiscovery}. The Markov Blanket of the monitoring target $y$ is used as the causal feature set.

\begin{equation}
\text{MB}(y) = \text{Parents}(y) \cup \text{Children}(y) \cup \text{Co-parents}(y).
\end{equation}

\subsubsection{Consensus Scoring}
Each method $m$ produces a score $S_m(f_i)$ that is min-max normalized to $\bar S_m(f_i) \in [0,1]$. The weighted consensus score is
\begin{equation}
C(f_i) = \sum_{m\in\mathcal{M}} w_m \cdot \bar S_m(f_i) \cdot \mathbb{I}(\bar S_m(f_i) > \tau_m),
\end{equation}
with a voting count $V(f_i) = \sum_m \mathbb{I}(\bar S_m(f_i) > \tau_m)$ and an agreement bonus
\begin{equation}
B(f_i) = \begin{cases} 0.15 & V(f_i)=3\\ 0.05 & V(f_i)=2\\ 0 & \text{otherwise}\end{cases}.
\end{equation}
A feature is selected if and only if
\begin{equation}
C(f_i) + B(f_i) \geq \tau_{\text{cons}} \ \wedge\ V(f_i) \geq V_{\min}.
\end{equation}
We use $\tau_{\text{cons}} = 0.30$, $\tau_m = 0.30$ for every $m$, and $V_{\min} = 2$, with category weights $w=(0.40, 0.30, 0.30)$ for tree-based, spectral, and causal criteria respectively.  

Let $f_i$ be a spurious feature and assume each method category independently evaluates $f_i$ with false-positive rate $\alpha_m = \mathbb{P}(\bar S_m(f_i)>\tau_m\mid\text{spurious})$. Under the consensus requirement $V(f_i)\geq V_{\min}=2$, the probability of selecting a spurious feature is bounded by
\begin{equation}
\mathbb{P}(\text{select }f_i\mid\text{spurious}) \leq \sum_{k=2}^{3}\binom{3}{k}\bar\alpha^k(1-\bar\alpha)^{3-k},
\end{equation}
where $\bar\alpha = \max_m \alpha_m$.
The three method categories operate on distinct principles, and their errors on spurious features are approximately independent. A confounder that reduces tree impurity is unlikely to also match spectral signatures and pass conditional independence tests. Consequently, the consensus mechanism achieves multiplicative suppression of spurious features relative to any single-criterion method.

\subsection{Recursive Lag-Feature Mechanism}
\label{sec:lag}
Industrial laboratory measurements are sampled far less frequently than process variables, yet they are precisely the ground-truth signal the model is trying to predict. Successive measurements of the same quality variable are also strongly autocorrelated, which makes the most recent laboratory value the single most informative input for predicting the next one. Failing to expose this signal to the model would discard the very information that distinguishes process monitoring from generic time-series forecasting.

We therefore augment the encoder embedding with a recursive lag vector before it enters the regression head. For each downstream query timestamp $t_q$, the lag-feature builder retrieves the $N_{\text{lag}}$ most recent laboratory measurements strictly before $t_q$ and emits the vector
\begin{equation}
\mathbf{l}(t_q) = \big[\,\tilde y_1, \Delta_1, m_1, \ldots, \tilde y_{N_{\text{lag}}}, \Delta_{N_{\text{lag}}}, m_{N_{\text{lag}}}\,\big] \in \mathbb{R}^{3 N_{\text{lag}}},
\end{equation}
where $\tilde y_i$ is the standardized $i$-th most recent lab value, $\Delta_i$ is the elapsed time in hours, and $m_i \in \{0,1\}$ is a presence mask that fires only when the lag is within the maximum-staleness budget $G_{\max}$. 

The mechanism is termed recursive because, when fresh laboratory measurements are temporarily unavailable, the model's own previous predictions are fed back into the lag builder as surrogate inputs, allowing continuous operation between sampling events. We refer to inference with real lab values as the closed-loop mode and inference with model-generated surrogates as the open-loop mode.

\subsection{Calibrated Uncertainty Head}
The adaptation head is a fully connected projection with two MC-dropout layers~\cite{gal2016dropout}. During inference, dropout remains active and $M$ stochastic forward passes are executed end-to-end, yielding an empirical distribution $\{\hat y_m\}_{m=1}^{M}$. The predictive mean and variance are
\begin{equation}
\mu_{\text{pred}}(\mathbf{x}) = \frac{1}{M}\sum_{m=1}^{M}\hat y_m,\qquad \sigma_{\text{pred}}^2(\mathbf{x}) = \frac{1}{M}\sum_{m=1}^{M}(\hat y_m - \mu_{\text{pred}})^2.
\end{equation}

In practice the raw MC-dropout variance often under- or over-estimates the true error magnitude, which leaves the resulting predictive intervals miscalibrated. We therefore introduce a scalar temperature $\gamma>0$ that rescales the predictive standard deviation, so that the $95\%$ predictive interval under a Gaussian approximation becomes
\begin{equation}
\text{PI}_{95} = \big[\mu_{\text{pred}} - 1.96\,\gamma\,\sigma_{\text{pred}},\ \mu_{\text{pred}} + 1.96\,\gamma\,\sigma_{\text{pred}}\big].
\end{equation}
The optimal temperature $\gamma^{*}$ is obtained by minimizing the Gaussian negative log-likelihood on a held-out calibration split, which admits the closed-form solution
\begin{equation}
\gamma^{*} = \sqrt{\frac{1}{N_{\text{cal}}}\sum_{j=1}^{N_{\text{cal}}} \frac{(y_j - \mu_j)^{2}}{\sigma_j^{2}}},
\end{equation}
where $\{(y_j,\mu_j,\sigma_j)\}_{j=1}^{N_{\text{cal}}}$ are the labels, predictive means, and predictive standard deviations on the calibration split. This adapts the temperature-scaling principle of~\cite{guo2017calibration} from classification to regression. The same dropout-based machinery extends to classification monitoring heads by applying MC-dropout to the softmax output.

For the soft sensing instantiation considered in this paper, the downstream loss combines prediction and regularization terms
\begin{equation}
\mathcal{L}_{\text{ft}}(\theta) = \frac{1}{N_{\text{down}}}\sum_{j=1}^{N_{\text{down}}}(\hat y_j - y_j)^2 + \lambda\|\theta_{\text{head}}\|_2^2,
\end{equation}
 When IPM-FM is adapted to classification-based monitoring tasks, the squared error is replaced by a cross-entropy term acting on the same backbone features.

\section{Experimental Validation}
\label{sec:experiments}

\subsection{Primary Downstream Task: Quality-Variable Soft Sensing}
Among the downstream tasks covered by IPM-FM, soft sensing of safety-critical quality variables is arguably the most label-scarce and most directly coupled to online control; it therefore serves as a stringent primary benchmark for the framework. 
The specific task chosen is diesel flash-point soft sensing at a commercial hydrotreating unit of the Parkland refinery.

\begin{figure}[t]
\centering
\includegraphics[width=0.999\columnwidth]{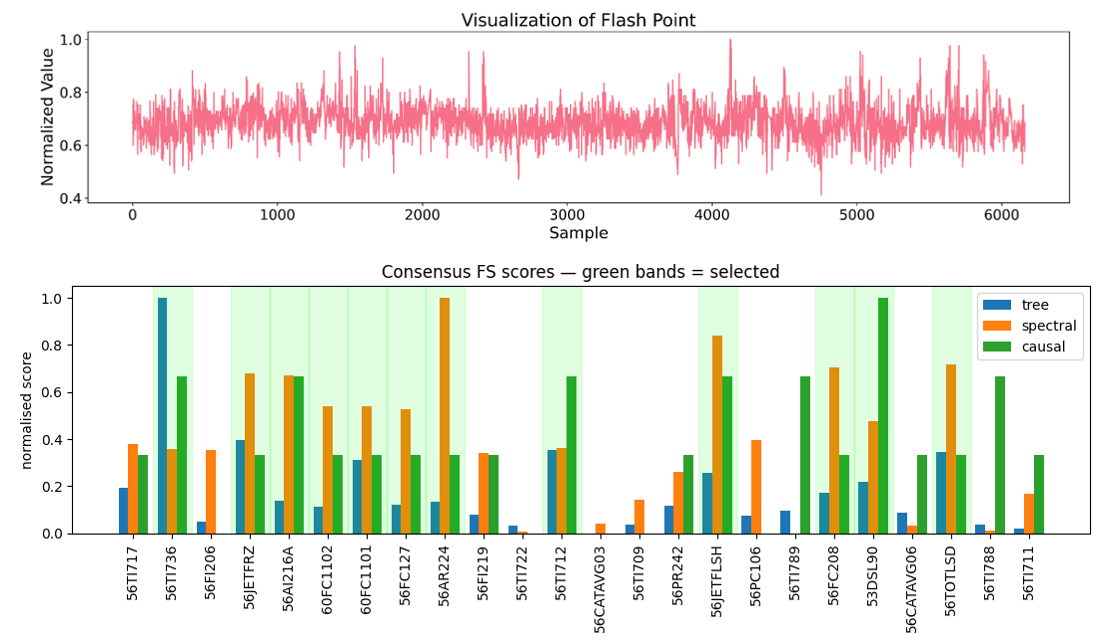}
\caption{Top panel: normalized flash-point trajectory over the seven-year dataset; bottom panel: consensus feature-selection scores per process tag, with green bands marking the selected channels.}
\label{fig:inputs}
\end{figure}

Flash point is a safety-critical property measured offline every 4--8 hours. We collected a seven-year dataset spanning June~2017 to June~2024, containing 6{,}157 quality samples. Each sample is aligned with 24 online process variables sampled every 10~min from the furnace, reactor, fractionator, and hydrogen system. Missing values were filled by interpolation, and all process variables were standardized. Figure~\ref{fig:inputs} shows the laboratory flash point together with a representative subset of selected process variables. We adopt a temporal split: the period from June~2017 to March~2024 is used for unlabeled pretraining and supervised fine-tuning, the period from April~2024 to June~2024 is used as the test set. Performance is reported as RMSE, MAE, MAPE, $R^2$, and 95\% predictive-interval coverage.

\subsection{Baselines}

We compare IPM-FM against two categories of baselines. 
The first category consists of classical machine learning methods, including decision tree, elastic net, orthogonal matching pursuit (OMP), partial least squares (PLS), $K$-neighbors, AdaBoost, random forest, gradient boosting, support vector machine (SVM) with an RBF kernel, LightGBM, extra trees, XGBoost, and CatBoost. 
The second category consists of sequence modeling baselines, including LSTM, Transformer, and Informer. 
The sequence baselines are run for 5 random seeds and we report mean ± standard deviation; classical learners are likewise run for 5 seeds where stochasticity exists.
All sequence baselines use the same input format and training budget as IPM-FM. In particular, the Informer baseline uses the same encoder backbone family as IPM-FM but is trained from scratch without consensus feature selection or self-supervised pretraining.

\subsection{Overall Performance}
\begin{figure}[t]
\centering
\includegraphics[width=0.99\columnwidth]{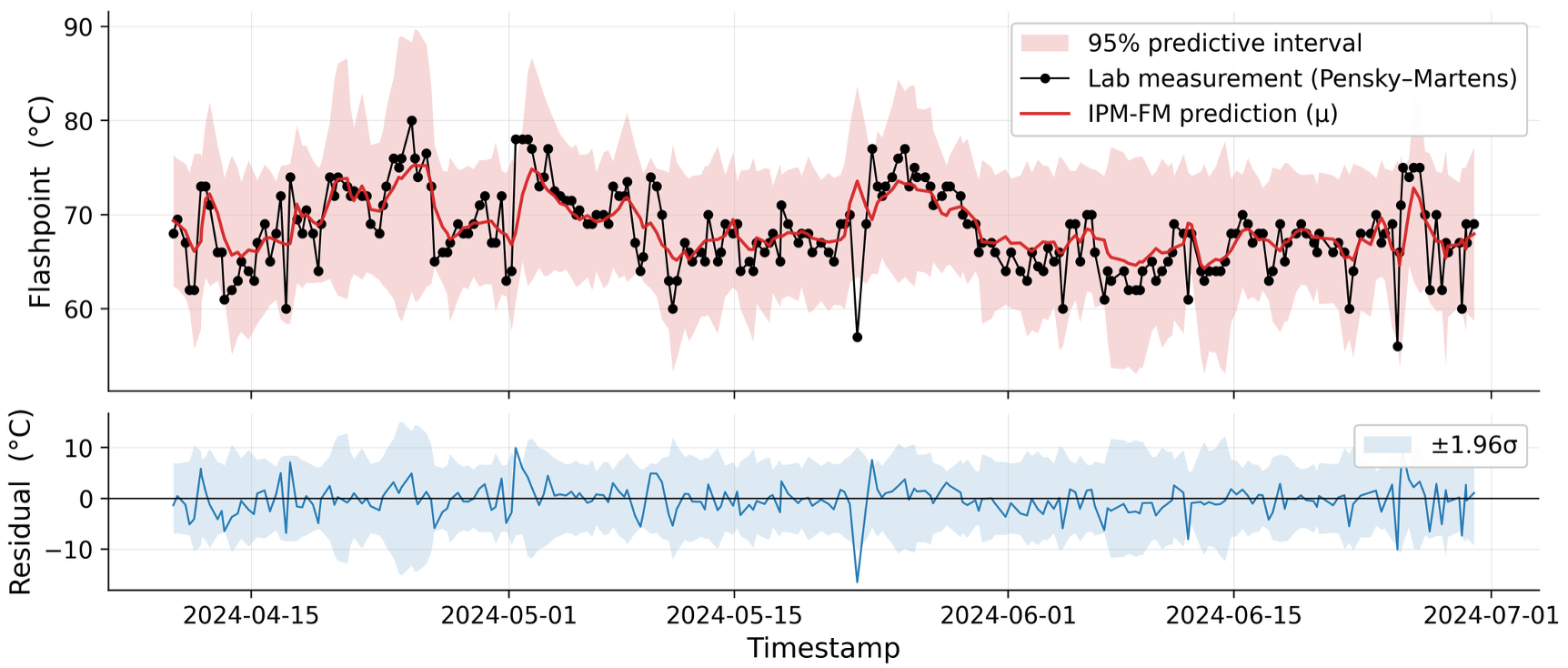}
\caption{IPM-FM flash-point prediction (red) with $\gamma$-calibrated 95\% predictive interval against laboratory measurements (black)}
\label{fig:prediction}
\end{figure}

Table~\ref{tab:main} summarizes the comparison across all baselines on the steady-state held-out window. IPM-FM achieves the lowest RMSE, MAE, and MAPE and the highest $R^2$ of any method tested. Compared to the strongest classical regressor (Partial Least Squares, RMSE~$3.26$), IPM-FM achieves an 8.3\% RMSE reduction and a 22\% increase in $R^2$. Compared to the strongest from-scratch sequence baseline (LSTM with no FS and no pretraining, RMSE~$3.50$), it delivers a 14.6\% RMSE reduction. Figure~\ref{fig:prediction} visualizes the IPM-FM mean prediction together with its $\gamma$-calibrated 95\% predictive interval and the per-sample residuals.

\begin{table*}[t]
\centering
 \caption{Flash-point soft sensing performance of IPM-FM against classical and from-scratch sequence baselines.}
\label{tab:main}
\footnotesize
\begin{tabular}{lcccc}
\toprule
Model & RMSE & MAE & MAPE(\%) & $R^2$ \\
\midrule
Decision Tree             & $6.54\pm0.14$ & $5.16\pm0.11$ & $7.66\pm0.16$ & $-1.36\pm0.10$ \\
Elastic Net               & $4.51\pm0.00$ & $3.71\pm0.00$ & $5.56\pm0.00$ & $-0.12\pm0.00$ \\
OMP                       & $4.47\pm0.00$ & $3.54\pm0.00$ & $5.27\pm0.00$ & $-0.10\pm0.00$ \\
Gradient Boosting         & $4.28\pm0.02$ & $3.09\pm0.01$ & $4.67\pm0.01$ & $-0.01\pm0.01$ \\
K-Neighbors               & $4.27\pm0.00$ & $3.38\pm0.00$ & $4.95\pm0.00$ & $-0.01\pm0.00$ \\
XGBoost                   & $4.23\pm0.00$ & $3.24\pm0.00$ & $4.88\pm0.00$ & $\phantom{-}0.01\pm0.00$ \\
LightGBM                  & $4.13\pm0.00$ & $3.17\pm0.00$ & $4.77\pm0.00$ & $\phantom{-}0.06\pm0.00$ \\
CatBoost                  & $4.08\pm0.10$ & $3.18\pm0.10$ & $4.80\pm0.15$ & $\phantom{-}0.08\pm0.05$ \\
AdaBoost                  & $3.91\pm0.03$ & $3.05\pm0.03$ & $4.51\pm0.05$ & $\phantom{-}0.15\pm0.01$ \\
Transformer     & $3.90\pm0.17$ & $3.14\pm0.16$ & $4.57\pm0.23$ & $\phantom{-}0.16\pm0.07$ \\
Extra Trees               & $3.76\pm0.02$ & $2.82\pm0.04$ & $4.24\pm0.05$ & $\phantom{-}0.22\pm0.01$ \\
Random Forest             & $3.75\pm0.03$ & $2.79\pm0.02$ & $4.21\pm0.03$ & $\phantom{-}0.22\pm0.01$ \\
SVM (RBF)                 & $3.72\pm0.00$ & $2.71\pm0.00$ & $4.03\pm0.00$ & $\phantom{-}0.24\pm0.00$ \\
Informer (scratch) & $3.68\pm0.18$ & $2.91\pm0.22$ & $4.24\pm0.30$ & $\phantom{-}0.25\pm0.07$ \\
LSTM            & $3.50\pm0.18$ & $2.73\pm0.16$ & $3.98\pm0.22$ & $\phantom{-}0.32\pm0.07$ \\
Partial Least Squares     & $3.26\pm0.00$ & $2.46\pm0.00$ & $3.65\pm0.00$ & $\phantom{-}0.41\pm0.00$ \\
\midrule
\textbf{IPM-FM (ours)}    & $\mathbf{2.99\pm0.03}$ & $\mathbf{2.14\pm0.03}$ & $\mathbf{3.21\pm0.04}$ & $\mathbf{0.50\pm0.01}$ \\
\bottomrule
\end{tabular}
\end{table*}

\begin{figure}[t]
\centering
\includegraphics[width=0.99\columnwidth]{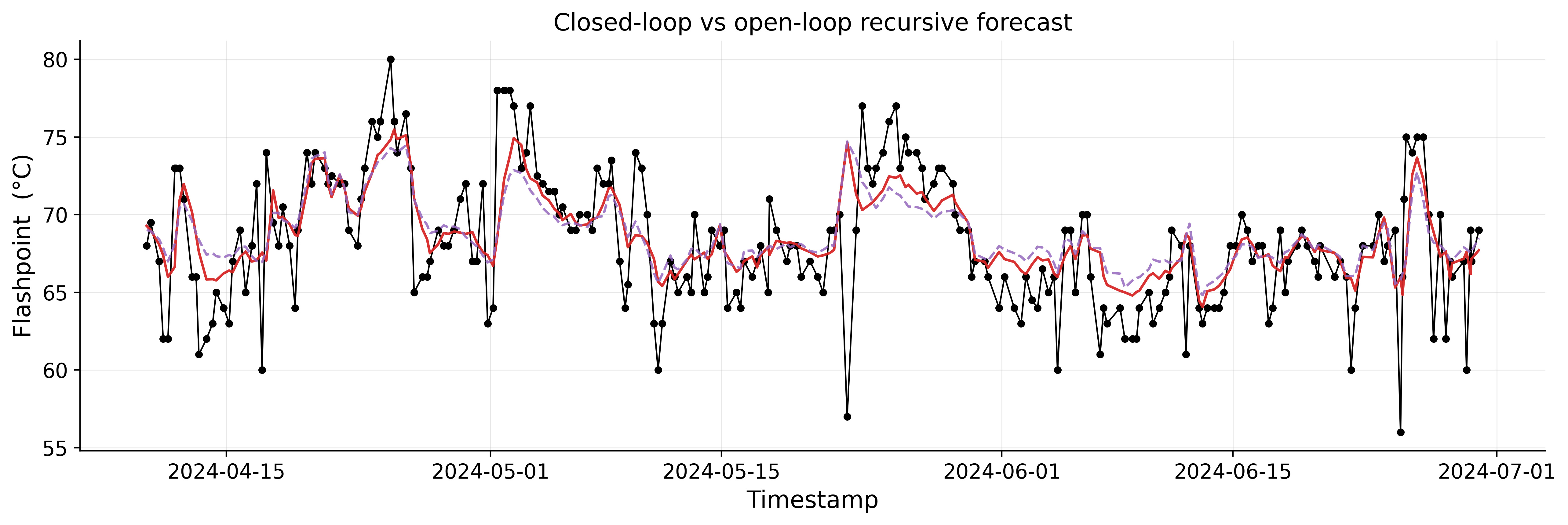}
\caption{Closed-loop (red) and open-loop recursive (purple) flash-point predictions on test set. }
\label{fig:closedopen}
\end{figure}

The recursive lag mechanism supports two evaluation regimes. Closed-loop inference uses the actual laboratory measurement as the lag input. Open-loop recursive inference uses only training labels to initialize the lag history and passes the model’s own predictions forward through the lag builder. Figure~\ref{fig:closedopen} compares the two regimes on the test window, showing that the open-loop mode tracks the laboratory trajectory with only a small RMSE penalty over closed-loop inference.

\subsection{Ablation Study}
We ablate each component of IPM-FM to isolate its individual contribution. Table~\ref{tab:ablation} reports the results. Removing the consensus feature selection causes the largest degradation (RMSE $2.99 \to 3.72$, $\Delta R^2 = -0.26$), confirming that multi-criteria channel selection is the primary driver of performance. Removing the recursive lag features causes the second-largest degradation (RMSE $2.99 \to 3.22$), demonstrating that the autoregressive coupling to prior laboratory values cannot be replaced by process-variable information alone. Removing self-supervised pretraining increases the RMSE to $3.11$, indicating that the masked reconstruction and operating-regime contrastive objectives provide transferable representations for downstream regression. 

Among the consensus-criterion ablations, removing the tree-based or spectral criterion degrades RMSE to $3.08$ and $3.09$ , respectively, while removing the causal criterion slightly improves RMSE to $2.94$. This suggests that tree-based and spectral evidence provide the strongest selection signals in this dataset, whereas the causal-discovery output is partly redundant with them for this particular soft-sensing task. We retain the causal criterion because it adds process-knowledge consistency and interpretability, even when its marginal effect on RMSE is dataset-dependent. Overall, the ablation results show that IPM-FM benefits primarily from consensus feature selection and recursive lag features, with self-supervised pretraining and MC-dropout calibration improving representation quality and uncertainty reliability, respectively.

\begin{table}[t]
\centering
\caption{Component-wise ablation of IPM-FM.}
\label{tab:ablation}
\footnotesize
\begin{tabular}{lccc}
\toprule
Variant & RMSE & MAE & $R^2$ \\
\midrule
Full IPM-FM                          & $\mathbf{2.99}$ & $\mathbf{2.14}$ & $\mathbf{0.50}$ \\
w/o causal criterion                 & $2.94$ & $2.11$ & $0.52$ \\
w/o tree-based criterion             & $3.08$ & $2.19$ & $0.48$ \\
w/o spectral criterion               & $3.09$ & $2.15$ & $0.47$ \\
w/o pretraining                      & $3.11$ & $2.23$ & $0.47$ \\
w/o recursive lag features           & $3.22$ & $2.42$ & $0.43$ \\
w/o consensus FS (full module)       & $3.72$ & $2.84$ & $0.24$ \\
\bottomrule
\end{tabular}
\end{table}

\section{Conclusion}
\label{sec:conclusion}
This paper addressed the label inefficiency and limited transferability of the prevailing one-task-one-model practice in industrial process monitoring by proposing IPM-FM, a foundation-model framework that decouples representation learning from task supervision. Experiments on a seven-year commercial hydrotreater dataset confirmed that the framework adapts to a label-scarce soft sensing task with calibrated predictions. IPM-FM attained an RMSE of 2.99 and an $R^2$ of 0.50, while holding 97\% empirical coverage on its 95\% predictive interval. It also outperformed the strongest classical and from-scratch sequence baselines by 8.3\% and 14.6\% RMSE. The methodological core consists of a consensus feature selector that suppresses spurious channels through cross-criterion agreement, a recursive lag mechanism that brings sparse laboratory values into inference, and a regression-side adaptation of temperature scaling for MC-dropout intervals. The unified pretraining--adaptation pipeline gives operating plants a route to reuse representation learning across the soft sensing, fault detection, and prognostic tasks they currently maintain in isolation.

\end{document}